\documentclass[runningheads,a4paper]{llncs}
\usepackage{graphicx}
\usepackage{cite}
\usepackage[T1]{fontenc}
\usepackage{balance}
\usepackage{amsmath}
\usepackage{amssymb,amsfonts}
\usepackage{algorithmic}
\usepackage[ruled]{algorithm2e}
\usepackage{graphicx}
\usepackage{textcomp}
\usepackage{xcolor, soul}
\usepackage{tabularx}
\usepackage{url}

\begin{document}
%
%\title{Semi-supervised learning optimization method based on BYOL algorithm}
\title{Integration of Self-Supervised BYOL in Semi-Supervised Medical Image Recognition}
\author{Hao Feng\inst{1} \and
Yuanzhe Jia\inst{1} \and
Ruijia Xu\inst{3} \and
Mukesh Prasad\inst{2} \and
Ali Anaissi\inst{1,2} \and
Ali Braytee\inst{2}}
\authorrunning{H. Feng et al.}
\institute{The University of Sydney, Camperdown, Australia \and
University of Technology Sydney, Ultimo, Australia \and
North University of China, China \\
%\email{ali.braytee@uts.edu.au}\\
}
\maketitle              % typeset the header of the contribution
\begin{abstract}

Image recognition techniques heavily rely on abundant labeled data, particularly in medical contexts. Addressing the challenges associated with obtaining labeled data has led to the prominence of self-supervised learning and semi-supervised learning, especially in scenarios with limited annotated data. In this paper, we proposed an innovative approach by integrating self-supervised learning into semi-supervised models to enhance medical image recognition. Our methodology commences with pre-training on unlabeled data utilizing the BYOL method. Subsequently, we merge pseudo-labeled and labeled datasets to construct a neural network classifier, refining it through iterative fine-tuning. Experimental results on three different datasets demonstrate that our approach optimally leverages unlabeled data, outperforming existing methods in terms of accuracy for medical image recognition.


\end{abstract}

\section{Introduction}

Artificial Intelligence (AI) has significant potential in medical applications but still faces challenges in medical image recognition~\cite{mlimarticle}. 
Medical image datasets are rich in spatial resolution, color channels, and diverse representations of organs, diseases, and anatomical structures. Obtaining relevant labels for medical classification is a primary hurdle, often requiring the expertise of medical professionals, leading to a time-consuming and logistically complex annotation process~\cite{litjens2017survey, braytee2022unsupervised}. 
Semi-supervised learning offers a potential solution for limited labeled data. This approach can improve model performance by incorporating additional unlabeled data. Similarly, self-supervised learning has emerged as a robust strategy for exploiting the inherent patterns in unlabeled data and acquiring representations that capture the underlying semantics.

We propose a method to address the challenge of limited labeled data in medical image classification by combining the advantages of self-supervised and semi-supervised learning. In our method, Bootstrap Your Own Latent (BYOL)~\cite{grill2020bootstrap}, a pre-training model for semi-supervised learning is employed to acquire useful representations from large amounts of unlabeled data. These representations are then fine-tuned using smaller labeled datasets to achieve high performance.
Our method offers two key advantages: firstly, BYOL enhances the model generalization by capturing structural and semantic information from unlabeled medical data. Secondly, semi-supervised learning optimizes the model performance by leveraging both labeled and unlabeled data.
%labeled data provides crucial supervision, while pseudo-labeling of unlabeled data enhances the training process and expands the available training examples.

\section{Related work}
Medical image recognition heavily relies on labeled datasets. Many AI techniques, such as deep convolutional neural networks, perform optimally when abundant data and high-quality annotations are available. Challenges persist, when faced with limited labeled data~\cite{esteva2017dermatologist}. To tackle this issue, unsupervised learning techniques like Sparse Coding (SC), Auto-encoders (AE), and Restricted Boltzmann Machines (RBMs) have been proposed. However, they are often limited to learning basic image features and struggle with the intricate characteristics present in medical images~\cite{ahn2020unsupervised, braytee2017multi}. Fortunately, leveraging unlabeled data through semi-supervised and self-supervised learning provides a promising avenue to reduce dependence on labeled data in medical image recognition.

Semi-supervised learning combines supervised and unsupervised methods, utilizing labeled and unlabeled data for enhanced performance.
When dealing with brain images that lack precise diagnostic information, semi-supervised support vector machines provide feature labels to overcome classification challenges~\cite{filipovych2011semisupervised}. 
%This approach is also effective in the context of imbalanced data. Multi-Curriculum Pseudo-Labelling (MCPL) improves the efficiency of semi-supervised learning, especially in the context of imbalanced medical image classification~\cite{peng2023faxmatch}.
This approach is also effective on imbalanced data. Multi-Curriculum Pseudo-Labelling (MCPL) evaluates the learning progress and adjusts the threshold for each category. This adaptation improves the efficiency of semi-supervised learning, especially for imbalanced medical image classification~\cite{peng2023faxmatch}.
Furthermore, semi-supervised learning can extract additional semantic information from unlabeled data~\cite{liu2020semisupervised}.
Overall, by leveraging unlabeled data, semi-supervised learning yields valuable information and training examples, thereby reducing labeling costs, addressing data imbalances, and enhancing model performance.

Although semi-supervised learning has reduced the expenses associated with medical image labeling, it still necessitates labeled data, a formidable challenge in medical images. There is a pressing need to diminish the model reliance on annotated data.
Recent studies have shown that incorporating self-supervised learning into the anomaly detection framework can significantly improve the accuracy by exploiting valuable information from the original unlabeled data~\cite{zhao2021anomaly}. 
%The feature acquisition capabilities of self-supervised learning are also widely used, with applications effectively realised in three common tasks: classification, positioning and segmentation~\cite{chen2019selfsupervised}. 
Moreover, a novel self-supervised learning algorithm, Bootstrap Your Own Latent (BYOL), was proposed and achieved a high classification accuracy on ImageNet, with the potential for further improvement using deeper architectures~\cite{grill2020bootstrap}. Unlike other self-supervised learning algorithms such as Contrastive Predictive Coding (CPC) and SimCLR, BYOL distinguishes itself by eliminating the need for generating negative samples for comparison. Therefore, the BYOL algorithm is a promising choice for augmenting semi-supervised models.

\section{A brief on BYOL}

BYOL is a self-supervised learning method for feature extraction in semi-supervised learning tasks. One of its key merits lies in its independence from labeled data, as it exclusively leverages unlabeled data during the training process. This approach enables the model to derive meaningful representations from a large volume of unlabeled data, thereby enhancing its efficacy in semi-supervised learning applications.

\begin{figure}[ht]
    \centering
    \includegraphics[width=1\textwidth,height=3cm]{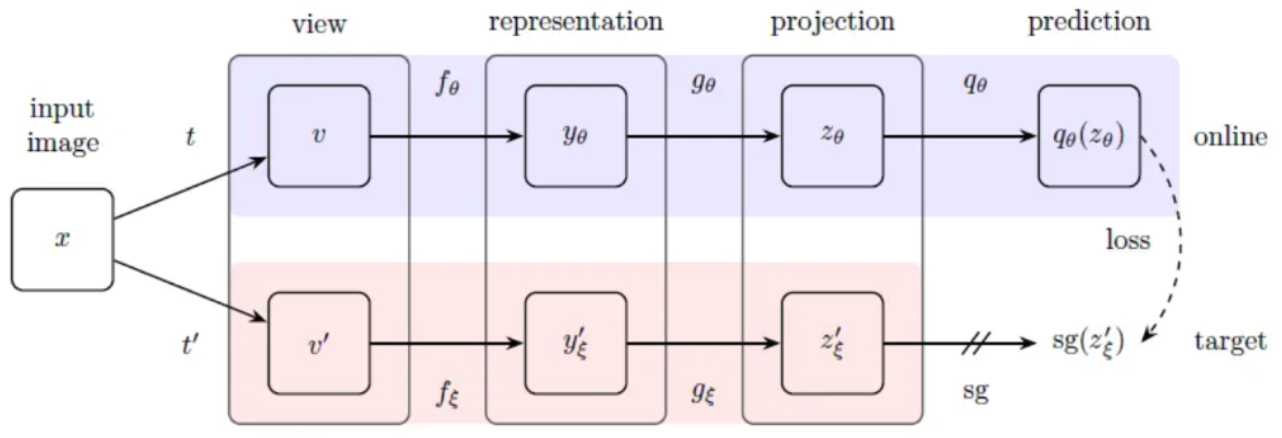}
    \caption{BYOL architecture.}
    \label{fig:byol}
\end{figure}

As shown in Fig.~\ref{fig:byol}, the core idea of BYOL is to develop two neural networks that learn from each other, the online network \(\theta\) and the target network \(\xi\), where $v$ is the input of \(\theta\) and $v'$ is the input of \(\xi\). 
$v$ and $v'$ are obtained by applying stochastic image augmentation methods $t$ and $t'$ on the input image $x$. 
$f_{\theta}$ and $f_{\xi}$ are encoders that share the same structure with different parameters so that $v$ and $v'$ get their respective corresponding representations $y_{\theta}$ and $y'_{\xi}$ after passing through $f_{\theta}$ and $f_{\xi}$, where $y_{\theta}$ is considered as a feature of the image that will be used eventually. 
After this, the online network \(\theta\) and the target network \(\xi\) map $y_{\theta}$ and $y'_{\xi}$ to the latent space to get $z_{\theta}$ and $z'_{\xi}$ using $g_{\theta}$ and $g'_{\xi}$, respectively.
The prediction result $q_{\theta}\left( z_{\theta} \right)$ can be obtained from the online network \(\theta\) by adding a layer of network structure $q_{\theta}$.
The target of BYOL is trying to make $q_{\theta}\left( z_{\theta} \right)$ as close to $z'_{\xi}$ as possible.

\begin{equation}
\mathcal{L}_{\theta,\xi} \triangleq \left\| {\overset{-}{q_{\theta}}~\left( z_{\theta} \right) - {\overset{-}{z}}_{\xi}^{'}} \right\|_{2}^{2} = 2 - 2 \bullet \frac{\left\langle {q_{\theta}\left( z_{\theta} \right),z_{\xi}^{'}} \right\rangle}{\left\| {q_{\theta}\left( z_{\theta} \right)} \right\|_{2} \bullet \left\| z_{\xi}^{'} \right\|_{2}}
\label{eq1}
\end{equation}

The loss function in Eq.~\ref{eq1}. But in real training processes, $v$ is also fed into the target network \(\xi\), while $v'$ is also fed into the online network \(\theta\). And these two are added together to get the final loss ${L}_{\theta,\xi}^{BYOL}$, as shown in Eq.~\ref{eq2}. Following this function, the target network \(\xi\) is updated according to the \(\theta\) obtained from the training process, thus achieving the effect that the online network is constantly approaching the target network.

\begin{equation}
{\mathcal{L}_{\theta,\xi}^{BYOL} = \mathcal{L}}_{\theta,\xi} + {\overset{\sim}{\mathcal{L}}}_{\theta,\xi}
\label{eq2}
\end{equation}

\section{Proposed method}

Our proposed method for medical image recognition is designed by utilising BYOL in a semi-supervised setting, which mainly consists of two steps: pre-training and fine-tuning.

\begin{figure}[ht]
    \centering
    \includegraphics[width=1\textwidth]{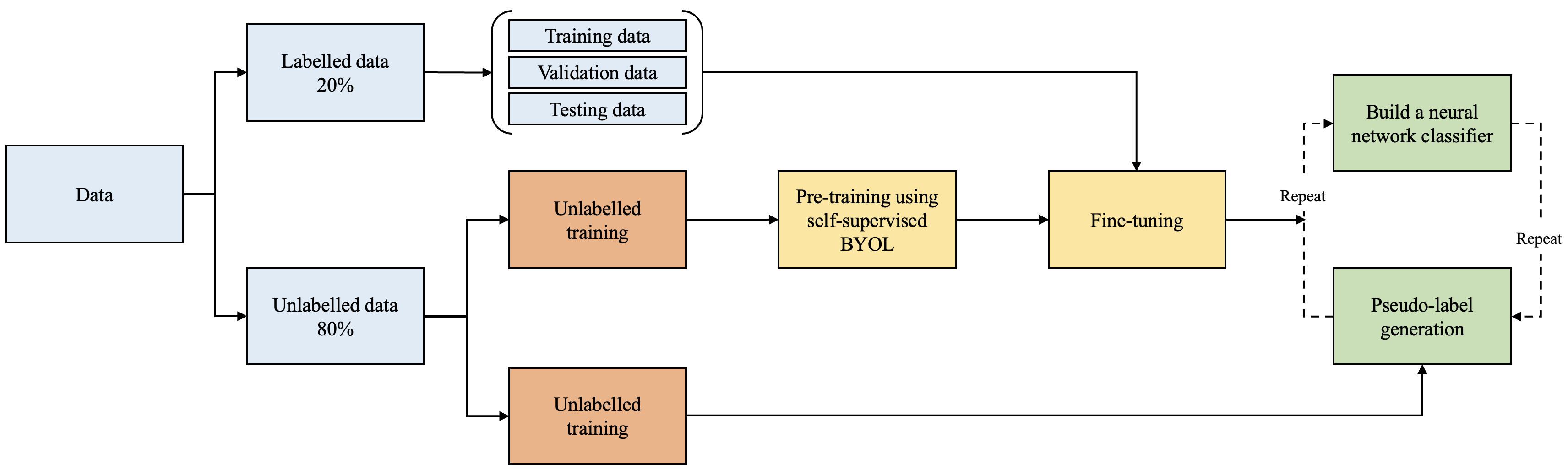}
    \caption{The proposed semi-supervised learning with BYOL.}
    \label{fig:mesh3}
\end{figure}

\subsection{Pre-training}

In our method, BYOL is employed to substitute the traditional pre-training process of semi-supervised learning, which relies on labeled data to generate the initial model. 
The optimisation process in Eq.~\ref{eq3}, where \(\eta\) is the learning rate. 
The loss $L(\theta, \xi)$ is calculated and optimized at each training step, while the network parameters \(\theta\) are dynamically updated by Stochastic Gradient Descent (SGD). The implementation of BYOL is relatively simple, with lower computing and memory requirements. As a result, we can streamline the initial phases of raw data preprocessing and feature extraction.

\begin{equation}
    \left. \theta\leftarrow optimizer\left( {\theta,\nabla_{\theta}\mathcal{L}_{\theta,\xi}^{BYOL},\eta} \right),~\xi\leftarrow\tau\xi + \left( {1 - \tau} \right)\theta, \right.
    \label{eq3}
\end{equation}

\subsection{Fine-tuning}

The feature extraction model generated by BYOL is fine-tuned with labeled data to construct a neural network classifier. This implementation is divided into three parts: constructing a neural network classifier, generating pseudo-labels, and model training.

\subsubsection{Constructing a neural network classifier}

The pre-trained BYOL model is fine-tuned using a smaller dataset chosen for image classification. This entails substituting the prediction layer with one customized for the target task and applying supervised learning on labeled data. 
The network parameters are initialized with the pre-trained weights from the prior task, capitalizing on weight inheritance to enhance the initial state, accelerate training convergence, and exploit features acquired by the pre-trained model. 
The prediction layer is crafted with input dimensions aligned with the output of the projection layer in the online network and is concurrently updated with the online network.
In addition, the loss function of BYOL transitions from image similarity to cross-entropy to provide intuitive insights and explanations. The cross-entropy loss function is defined in Eq.~\ref{eq:entropy}, where $y$ represents the one-hot encoded ground truth labels for $N$ samples, and $p$ denotes the predicted class probabilities.

\begin{equation}
H(y, p) = -\sum_{i=1}^{N} y_i \log\left(\frac{e^{p_i}}{\sum_{j=1}^{N} e^{p_j}}\right)
\label{eq:entropy}
\end{equation}

\subsubsection{Generating pseudo-labels}

Pseudo-labels are formulated by leveraging unlabeled data and generating predictions using the pre-trained model. These synthetic labels are then integrated into the original dataset, effectively updating the sample labels within the unlabeled data. Subsequently, the predicted labels are appended to the pseudo-label repository of the neural network classifier. This combination of pseudo-labels with a limited set of labeled data enlarges the dataset, encompassing both authentic and synthetic labels, albeit with potential noise. This expanded training dataset introduces a broader spectrum of data distributions, increased sample diversity, and enhanced informational content. Consequently, it empowers the model to acquire a more extensive understanding, elevating its capacity for generalization.

\subsubsection{Model training}

The modified BYOL model undergoes retraining, leveraging both labeled and pseudo-labeled data. This fusion facilitates improved generalization and heightened accuracy in addressing the target task. Each iteration involves fine-tuning the model with labeled data, generating pseudo-labeled data based on the current model, and retraining the model with a combined dataset comprising both labeled and pseudo-labeled instances. This iterative optimization consistently elevates the model performance, enhancing its resilience and generalization capabilities. Striking a balance between the objectives of model improvement is also crucial throughout this process. Determining the optimal point for iterations, guided by specific requirements and constraints, is essential in selecting the most effective model.

\section{Experiments and results}

\subsection{Datasets}

Three datasets are used in our experiments: OCT2017~\cite{kermany2018labeled}; COVID-19 X-ray\cite{sahoo2022potential}; and Kvasir\cite{liu2023semisupervised}. 
OCT2017 consists of 109,312 images, of which 108,312 are used for training, 32 for validation and 968 for testing. 
COVID-19 X-ray contains 2,905 posterior-to-anterior chest X-ray images, including positive cases, viral pneumonia cases and normal cases.
Kvasir collects manually annotated images covering several classes, including anatomical landmarks, pathological findings and endoscopic procedures, each class consisting of hundreds of images.

\subsection{Hyperparameter tuning}
Due to space constraints, we only demonstrate the tuning process on the OCT2017 dataset. Three hyperparameters are tested: the number of epochs, the learning rate, and the number of pseudo-labels. We first searched for the best combination of epochs and learning rates, exploring a range of epochs from 50 to 250 and a range of learning rates from 0.001 to 0.03. We employ accuracy as an evaluation metric. The model achieved its highest accuracy of 0.95 with 250 epochs and a learning rate of 0.001, although the computation time increased with a larger epoch count (see Table~\ref{tab:table1}). We then searched for the optimal number of pseudo-labels using grid search with values of 500, 1000, and 2,000, which attained accuracy scores of 0.961, 0.962, and 0.966, respectively. This allowed us to establish the optimal hyperparameters for our model: 250 epochs, a learning rate of 0.001, and 2,000 pseudo-labels.

%\begin{figure}[h]
%    \centering
%    \includegraphics[width=0.5\textwidth, height=5cm]{param1.png}
%    \caption{Selecting the optimal number of epochs using losses on OCT2017}
%    \label{fig:param1}
%\end{figure}

\begin{table}[ht]
\centering
\caption{Selecting the optimal epoch and learning rate on OCT2017.}
\label{tab:table1}
\begin{tabular}{l|c|c|c}
 Learning Rate/Epoch & 50 & 150 & 250\\ \hline
 0.03  & 0.51  & 0.86  & 0.83 \\
 0.01  & 0.62  & 0.81 & 0.90\\
 0.001  & 0.83  & 0.93  & \textbf{0.95}\\ \hline
 Computation Time  & 15 min  & 42 min  & 94 min\\
\end{tabular}
\end{table}

%\begin{table}[ht]
%\centering
%\caption{Selecting the optimal number of pseudo labels on OCT2017, while the learning rate %is 0.001 and the number of epochs is 250.}
%\label{tab:table2}
%\begin{tabular}{c|c}
%  Pseudo Label & Accuracy \\ \hline
% 500  & 0.961 \\
% 1000  & 0.962\\
% 2000  & \textbf{0.966}\\
%\end{tabular}
%\end{table}

\subsection{Results}

\begin{table}[ht]
\centering
\caption{Classification results using accuracy on three datasets.}
\label{tab:performance}
\begin{tabular}{l|c|c|c}
\textbf{Method/Dataset} & \textbf{OCT2017} & \textbf{COVID-19} & \textbf{Kvasir} \\
\hline
%PL \cite{lee2013pseudo} & 0.91 & 0.92 & 0.91 \\
MeanTeacher \cite{tarvainen2017mean} & 0.92 & 0.93 & 0.92 \\
MixMatch \cite{berthelot2019mixmatch} & 0.93 & 0.94 & 0.91 \\
FixMatch \cite{sohn2020fixmatch} & 0.92 & 0.93 & 0.92 \\
VAT \cite{miyato2018virtual} & 0.92 & 0.91 & 0.91 \\
VATNM \cite{cui2020towards} & 0.93 & 0.92 & 0.92 \\
GLM \cite{gyawali2020semi} & 0.94 & 0.93 & 0.93 \\
VTS \cite{wang2021deep} & 0.95 & 0.95 & 0.93 \\
NNM \cite{liu2023semi} & 0.95 & 0.96 & 0.93 \\
Ours & \textbf{0.966} & \textbf{0.987} & \textbf{0.976} \\
\end{tabular}
\end{table}

The experiments aim to evaluate the model performance across various datasets, including OCT2017, COVID-19 X-ray and Kvarsir\footnote{We will open source the code for reproducibility: GitHub link}.
%The proposed model is built using optimal hyper-parameters derived from the previous section. 
%We will open source the code later to ensure the reproducibility of our experiments\footnote{GitHub link will appear here}.
Table~\ref{tab:performance} displays a comparative analysis of results across three distinct datasets, assessing the proposed model against various semi-supervised counterparts such as MeanTeacher, MixMatch, FixMatch, VAT, VATNM, GLM, VTS and NNM. 
Evaluation metrics focus on test set accuracy. 
For OCT2017, benchmark models achieved scores between 0.92 and 0.95. Similarly, accuracy scores fell within the 0.91 to 0.96 on COVID-19 X-ray, and maintain performance within the 0.91 to 0.93 range on Kvasir.
Noteworthy is the exceptional performance of our proposed method across all three datasets, achieving scores of 0.966 for OCT2017, 0.987 for COVID-19, and 0.976 for Kvasir. These results underscore the effectiveness and competitiveness of our approach when compared to existing methods.

\section{Conclusion}

This paper introduces a method to improve the performance of semi-supervised learning on medical image data by integrating self-supervised BYOL. Experimental findings demonstrate that BYOL markedly boosts semi-supervised learning, diminishing the dependence on labeled data and setting it apart from existing semi-supervised approaches. 
%Across three medical image datasets, the proposed method surpasses other semi-supervised approaches.

\bibliographystyle{splncs04_unsort}
\balance
\bibliography{sample-base}

\end{document}